\documentclass[12pt]{custarticle}

\usepackage{hyperref}
\usepackage{float}
\usepackage{verbatim} 
\usepackage{apalike}
\usepackage{mathtools} 
\usepackage{booktabs} 
\usepackage{tikz} 
\usepackage{color}
\usepackage{bbm}
\usepackage{geometry}
\usepackage{setspace}
\usepackage{graphicx}	
\usepackage{algorithm}
\usepackage{multicol}
\usepackage{amsthm}
\usepackage{amssymb}
\usepackage{mathtools}
\usepackage{subcaption}

\newcommand{\rev}{R}
\makeatletter
\def\ps@pprintTitle{%
  \let\@oddhead\@empty
  \let\@evenhead\@empty
  \let\@oddfoot\@empty
  \let\@evenfoot\@oddfoot
}
\makeatother

\newcommand{\name}{\textit{InteraSSort}}
\biboptions{authoryear}

\begin{document}
\begin{frontmatter}

\title{\name{}: Interactive Assortment Planning Using Large Language Models}

\author[label1]{Saketh Reddy Karra\corref{cor1}}
\ead{skarra7@uic.edu}

\author[label1]{Theja Tulabandhula}
\ead{theja@uic.edu}

\address[label1]{University of Illinois Chicago, 601 S Morgan St, Chicago, IL 60607, United States}

\begin{abstract}
 Assortment planning, integral to multiple commercial offerings, is a key problem studied in e-commerce and retail settings. Numerous variants of the problem along with their integration into business solutions have been thoroughly investigated in the existing literature. However, the nuanced complexities of in-store planning and a lack of optimization proficiency among store planners with strong domain expertise remain largely overlooked. These challenges frequently necessitate collaborative efforts with multiple stakeholders which often lead to prolonged decision-making processes and significant delays. To mitigate these challenges and capitalize on the advancements of Large Language Models (LLMs), we propose an interactive assortment planning framework, \name{} that augments LLMs with optimization tools to assist store planners in making decisions through interactive conversations. Specifically, we develop a solution featuring a user-friendly interface that enables users to express their optimization objectives as input text prompts to \name{} and receive tailored optimized solutions as output. Our framework extends beyond basic functionality by enabling the inclusion of additional constraints through interactive conversation, facilitating precise and highly customized decision-making. Extensive experiments demonstrate the effectiveness of our framework and potential extensions to a broad range of operations management challenges.

\end{abstract}


\end{frontmatter}

\section{Introduction}
\begin{figure}[htbp]
  \centering
  \includegraphics[width=0.6\linewidth]{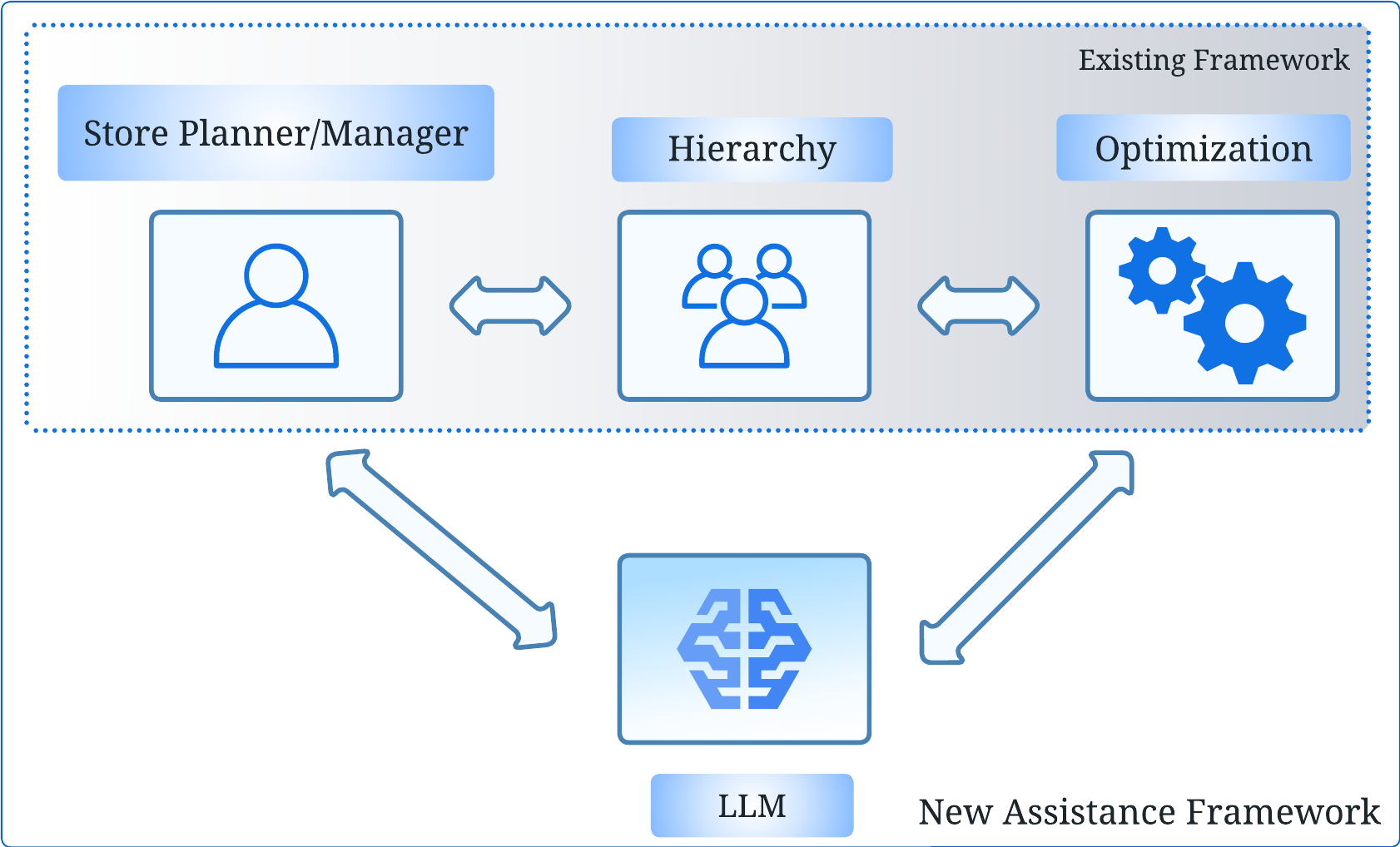}
  \caption{Incorporating LLM as an intelligent assistant to the existing framework.}
  \label{fig:chat}
\end{figure}

    Assortment planning~\citep{kok2015assortment, rossi2003bayesian} is a pivotal marketing strategy employed by managers/store planners in the retail industry. These planners are responsible for designing the layout and product assortments for physical retail stores to maximize sales, customer satisfaction, and profitability. These seasoned planners, with their deep domain knowledge, often need to generate insights for questions that involve variations of the assortment optimization problem. However, due to the complexity inherent in store planning, as well as the absence of optimization expertise among store planners, significant challenges often arise. The process of insight generation as a result requires collaboration with multiple professionals, resulting in prolonged decision-making processes and significant delays. Consequently, there is a clear demand for a framework designed to assist store planners as shown in Figure~\ref{fig:chat}. This framework should be able to provide dynamic solutions to various assortment planning problems, all while eliminating the necessity for a detailed understanding of technical optimization framework, thereby assisting in the decision-making process.
    
    The advent of artificial intelligence (AI) has brought about a revolutionary transformation in the way businesses operate. Among these cutting-edge innovations, large language models (LLMs) such as GPT-4~\cite{openai2023gpt4} and LLaMA~\cite{touvron2023llama} have emerged as pioneers of generative AI, leading the forefront of the latest technological disruptions. However, it is only in recent years that the intersection of AI and marketing has captured the attention of researchers. This has prompted further investigations into AI-related topics and their roles in marketing~\cite{jain2023prospects}. In light of this, LLMs with their advanced capabilities can serve as a fundamental component in creating an interactive framework tailored for solving marketing challenges, such as assortment optimization, thus assisting the store planners in making informed decisions.
 
    LLMs hold immense potential as general task solvers; recent advancements have pushed their functionality beyond mere chatbot capabilities, positioning them as assistants or even replacements for domain experts. However, employing LLMs directly to solve intricate assortment optimization problems presents numerous challenges due to the diverse input data formats and the inherently combinatorial nature of the complex optimization problem at hand. Furthermore, despite the impressive capabilities of LLMs, they encounter difficulties when confronted with complex reasoning tasks that demand specialized functionalities, such as arithmetic calculations~\cite{frieder2023mathematical} and information retrieval~\cite{li2023evaluating}. Moreover, LLMs lack the ability to solve simple optimization problems independently, necessitating the integration of solvers like the \emph{Cplex} and \emph{Gurobi}~\citep{anand2017comparative}. 

    In addition to integrating LLMs, a significant challenge lies in selecting suitable assortment optimization algorithms capable of delivering swift and scalable solutions, keeping the aspect of interactivity at the forefront. Addressing these challenges, we propose a collaborative framework \name{} that effectively integrates LLMs with optimization tools to tackle the assortment planning problem in an interactive manner. \name{} enables planners to present their optimization objectives using natural language through input prompts, and the framework will respond by making appropriate calls to optimization tools and solvers. Our approach goes beyond basic functionality by incorporating the ability to include additional constraints through text prompts and generate solutions interactively. We summarize the list of our contributions below.

\begin{itemize}
    \item We design \name{} to feature a user-centric chat interface via Streamlit~\footnote{\url{https://streamlit.io/}}, with LLMs and optimization algorithms seamlessly integrated into the backend to carry out the tasks based on the input prompts provided by the user.
    
    \item Our framework leverages the conversational history and function-calling capability of LLMs to accurately invoke the requisite functions in response to input prompts, facilitating the execution of optimization scripts to deliver solutions to the user.
\end{itemize}

\section{Related Work} \label{sec:rw}
In this study, we extend upon two key streams of research: (a) AI applications in marketing and (b) Language model tools. We briefly discuss some of the related works below.

\paragraph{Applications of AI in Marketing}
~\cite{verma2021artificial} explored the role of AI and disruptive technologies in business operations, explicitly highlighting the use of chatbots and language models to enhance the customer experience and customer relationship management systems. Similarly, ~\cite{de2022machine} presented a comprehensive taxonomy of machine learning and AI applications in marketing, emphasizing customer-facing improvements such as personalization, communication, recommendations, and assortments, as well as the benefits of machine learning on the business side, including market understanding and customer sense. In their literature review, ~\cite{duarte2022machine} identified recommender systems and text analysis as promising areas for chatbot utilization in marketing. ~\cite{fraiwan2023review} discussed the applications, limitations, and future research directions pertaining to advanced language models in marketing. Building on earlier works, we explore the application of AI to solve assortment planning problem using LLMs.

\paragraph{Tools and their integration with LLMs}:
Researchers have made significant strides in using LLMs to tackle complex tasks by extending their capabilities to include planning and API selection for tool utilization. For instance, \cite{schick2023toolformer} introduced the pioneering work of incorporating external API tags into text sequences, enabling LLMs to access external tools. TaskMatrix.AI \cite{liang2023taskmatrix} utilizes LLMs to generate high-level solution outlines tailored to specific tasks, matching subtasks with suitable off-the-shelf models or systems. HuggingGPT \cite{shen2023hugginggpt} harnesses LLMs as controllers to effectively manage existing domain models for intricate tasks. Lastly, \cite{qin2023tool} proposed a tool-augmented LLM framework that dynamically adjusts execution plans, empowering LLMs to proficiently complete each subtask using appropriate tools. \cite{li2023large} introduced the Optiguide framework, leveraging LLMs to elucidate supply chain optimization solutions and address what-if scenarios. In contrast to the aforementioned approaches, \name{} harnesses the power of LLMs to enable interactive optimization in the context of assortment planning.

\section{Preliminaries} \label{sec:plm}
In this section, we discuss the assortment planning problem in detail and key questions that can be answered through interactivity.

\subsection{Assortment planning}
The assortment planning problem involves choosing an assortment among a set of feasible assortments ($\mathcal{S}$) that maximizes the expected revenue. Consider a set of products indexed from $1$ to $n$ with their respective prices being $p_1, p_2, \cdots p_n$. The revenue of the assortment is given by $\rev(S) = \sum_{k \in S} p_k \times \mathbb{P}(k|S)$ where $S \subseteq \{1,...,n\}$. The expected revenue maximization problem is simply: $\max_{S \in \mathcal{S}} \rev(S)$. Here $\mathbb{P}(k|S)$ represents the probability that a user chooses product $k$ from an assortment $S$ and is determined by a choice model. 

The complex nature of the assortment planning problem requires the development of robust optimization methodologies that can work well with different types of constraints and produce viable solutions within reasonable time frames. In this study, we adopt a series of scalable efficient algorithms~\citep{tulabandhula2022optimizing} addressing the assortment optimization problem.

\subsection{Key questions answered through interactivity}
As previously highlighted, store planners with their deep domain knowledge, often need to generate insights for questions that involve variations of the assortment optimization problem. Accordingly, our framework \name{} needs to interactively address the key questions outlined below.

\begin{itemize}
    \item What would be the optimal assortment when constrained by a specific limit on the assortment size?
    \item What constitutes the optimal assortment when a product cannot be included?
    \item What is the expected revenue of the assortment if a product is to be part of the selection? 
\end{itemize}

\section{The \name{} Framework} \label{sec:framework}
Solving an assortment planning problem in real-world scenarios involves several crucial steps. The process begins with exhaustive data collection and analysis, followed by selecting a suitable choice model and estimating its parameters. Subsequently, the relevant optimization algorithm is executed to determine the optimal assortment. The process concludes with communication and implementation of derived decisions among various stakeholders. 

\begin{figure}[htbp]
  \centering
  \includegraphics[width=1\linewidth]{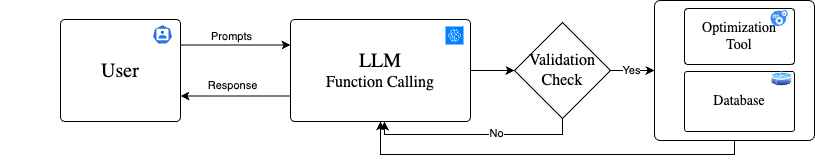}
  \caption{Overview of \name{} framework.}
  \label{fig:workflow}
\end{figure}

Our framework, \name{}, as shown in Figure~\ref{fig:workflow} takes input via user prompts. The LLM with function calling ability translates these prompts into the desired format and executes the optimization tools following a validation check. The generated solutions are relayed back to the user via the LLM. This interactive process repeats as the user provides additional prompts, fostering a dynamic exchange of information. The process discussed above is structured into multiple stages: 1) prompt design, 2) prompt decomposition, and 3) tool execution \& response generation.

\subsection{Prompt design}
\name{} uses an LLM to perform detailed analyses of user requests, which are submitted as text prompts. Therefore, the design of the prompts is crucial for accurately capturing and utilizing user requests in later stages. To facilitate this, the framework requires a standardized template for input prompts to systematically extract constraints and other relevant information i.e., the dataset and choice model to be used.

\subsection{Prompt decompostion}
In this stage, \name{} leverages the function-calling capabilities of the LLM to break down the standardized prompts. This feature empowers the LLM to generate JSON objects containing arguments for calling functions that conform to the predefined specifications required for solving the optimization problem. The function calling template incorporates multiple slots, such as `model', `dataset', and `cardinality', to represent various variables and constraints as shown in Figure. By adhering to these task specifications, \name{} efficiently utilizes the LLM to analyze user requests and accurately parse them.

To facilitate interactive multi-turn conversations, \name{} has the capability to append chat history to the follow-up prompts. This is crucial, as these prompts may lack the entire context required to generate a solution. Consequently, whenever a user poses a follow-up question,  \name{} can reference past interactions and trace prior user responses to answer subsequent questions. This functionality enables \name{} to more effectively manage context and respond to user requests in multi-turn dialogues.

\begin{figure}[htbp]
  \centering
  \includegraphics[width=0.8\linewidth]{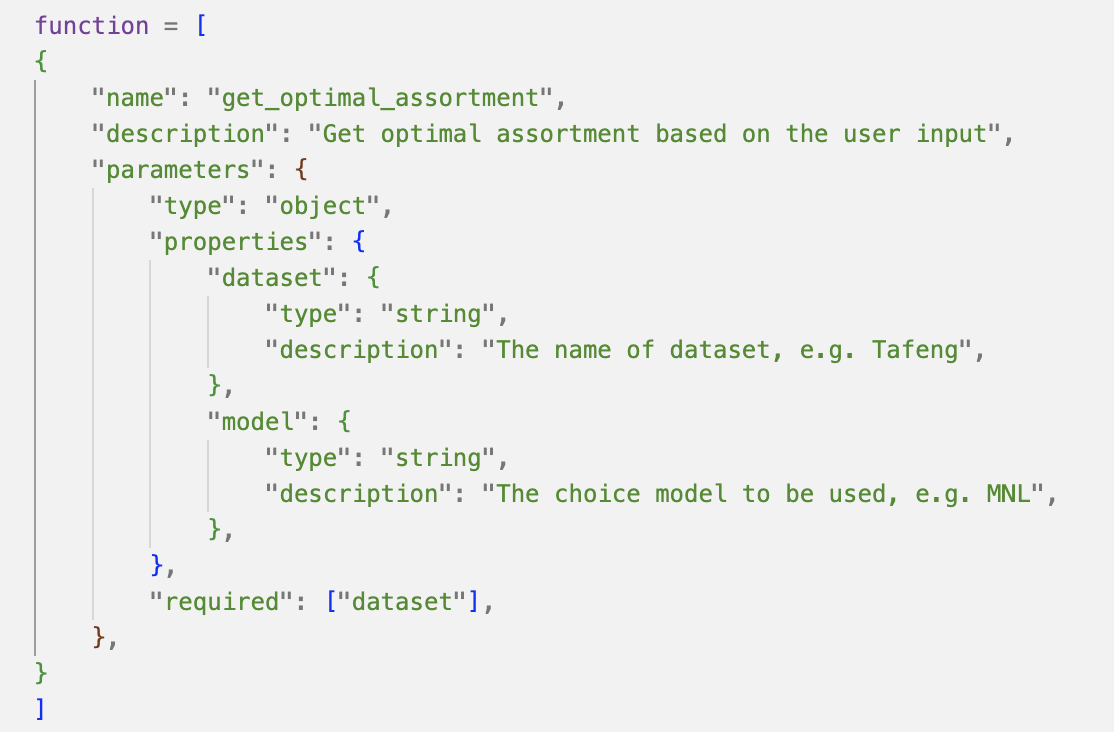}
  \caption{Potential function configuration for prompt decomposition.}
  \label{fig:function }
\end{figure}

\subsection{Tool execution \& response generation}
\name{} effectively manages and processes the output received from the prompt decomposition stage. This involves conducting thorough validation checks, such as range and consistency assessments, to ensure the accuracy and reliability of the decomposed prompts. \name{} maintains a comprehensive database for parameters of choice models across multiple datasets. Upon successful validation, it retrieves corresponding parameters based on the choice model identified in the decomposed prompt. Utilizing the choice model parameters and any other constraints as arguments, \name{} executes the optimization scripts using tools like optimization solvers to achieve the best possible results. Finally, \name{} enables the LLM to receive these results as input and generate responses in user-friendly language.

\section{Illustration} \label{sec:illustration}
In this section, we discuss the components needed to run our experiments followed by an illustrative example.

\subsection{Ta-Feng dataset}
Ta-Feng~\footnote{\url{https://www.kaggle.com/datasets/chiranjivdas09/ta-feng-grocery-dataset}} is a grocery shopping dataset released by ACM RecSys. The dataset contains detailed transactional records of users over a period of $4$ months, from November $2000$ to February $2001$. The total number of transactions in this dataset is $817,741$ which are associated with $32,266$ users and $23,812$ products.

\subsection{Multinomial logit (MNL)}
The MNL model~\citep{luce2012individual} is one of the most extensively studied discrete choice models and is frequently utilized across various marketing applications. The parameters of the MNL model are represented by a vector $\mathbf{v} = \left(v_0, v_1, \cdots v_{n}\right)$ with $0 \leq v_i \leq 1 \;\;\forall i$. Parameter $v_i, \ 1\leq i \leq n $, captures the preference of the user for purchasing product $i$. Under this model, the probability that a user chooses product $k$ from an assortment $S$ is given by $\mathbb{P}(k|S) = v_l/(v_0 + \sum_{k' \in S} v_{k'})$.

\subsection{LLM}
We employ the \texttt{gpt-3.5-turbo} variant from the GPT model series as our primary LLM. The model is publicly accessible through the OpenAI API~\footnote{\url{https://platform.openai.com/}}.

\begin{figure}[htbp]
\captionsetup[subfigure]{}
\begin{multicols}{2}
    \includegraphics[width=1.1\linewidth]{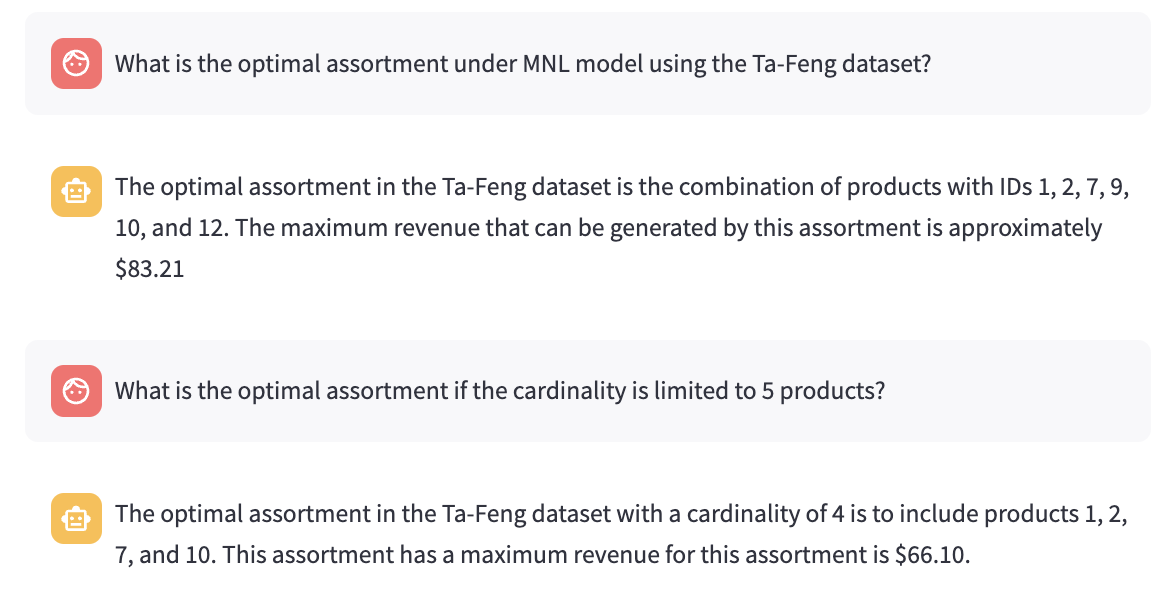}\par
    \includegraphics[width=1.1\linewidth]{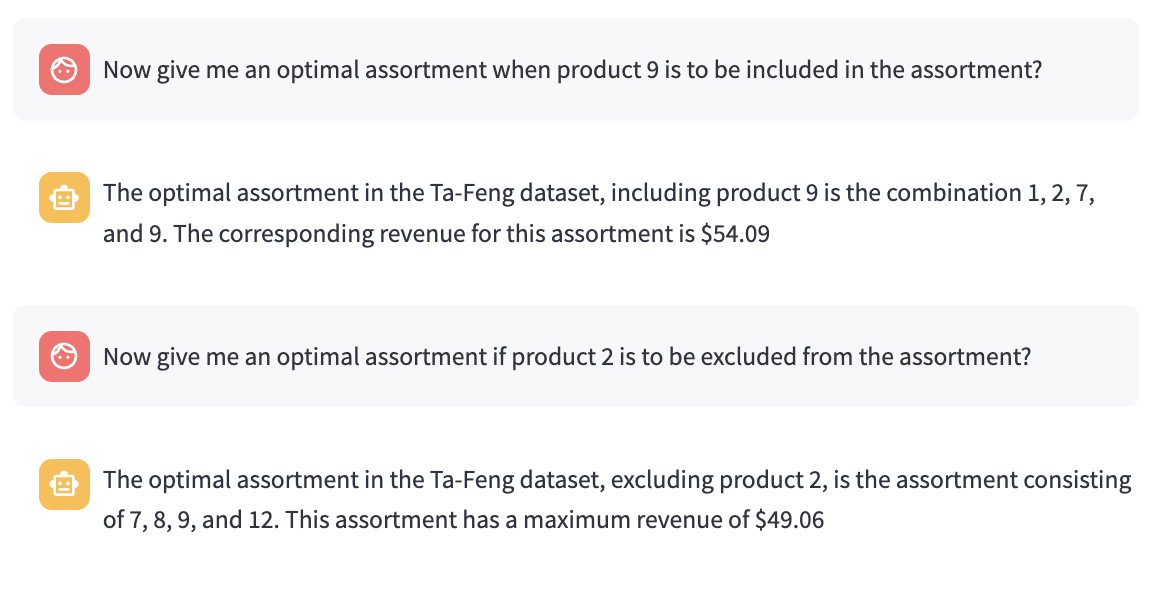}\par
\end{multicols}
\caption{Illustrative example showing the interactions with \name{} framework.}
\label{fig:dist_comp}
\end{figure}

\subsection{Illustrative example}
We demonstrate the data flow in the \name{} framework using a user question: \textit{`What is the optimal assortment for the Ta-Feng Dataset using the MNL model?'}, as shown in Figure~\ref{fig:dist_comp}. The question is entered as an input prompt via user interface. Utilizing the LLM's function-calling capability, the input is parsed (i.e., identifying `Ta-Feng' as the dataset and `MNL' as the choice model). Based on the parsed input, \name{} efficiently leverages parameter data for the Ta-Feng dataset and invokes the relevant function. This function then processes the arguments, executes the MNL optimization script, and communicates the outcomes through the interface. Whenever a user poses a follow-up question in the form of any additional constraint, such as \textit{`I want an optimal assortment where assortment size is limited to 5 products'}, the system makes use of the decomposed inputs from the previous interaction, along with the constraint limiting the size of optimal assortment and passes these as arguments to the relevant function. This function then reruns the optimization script, with the set of updated arguments and returns the solution, to the LLM which communicates to the user.

\section{Conclusion} \label{sec:conclusion}
In this paper, we introduced \name{}, an interactive framework designed to empower planners with limited optimization expertise in deriving insightful solutions to the assortment planning problem. \name{} facilitates interactive optimization by generating responses to variations of the optimization problem based on user requests. By harnessing the inherent strengths of instruction-tuned LLMs such as comprehension and reasoning, \name{} excels in interpreting user requests and breaking them down into distinct function parameters, that enable flexible assortment planning. Subsequently, \name{} intelligently calls and executes the most appropriate optimization tools and translates the solutions into concise, easily interpretable responses for the user. Overall, \name{} enables working with assortment planning problem effectively through interaction, and the framework can be easily extended to other marketing problems in the field of operations management.

\bibliography{references}

\end{document}